\documentclass{article}
\usepackage{spconf,amsmath,graphicx,hyperref}
\usepackage{algorithm,algorithmic,booktabs}

\title{TriCon-Fair: Triplet Contrastive Learning for Mitigating Social Bias in Pre-trained Language Models}

\name{Chong Lyu$^{\dagger}$, Lin Li$^{\dagger\star}$\thanks{$\star$ Corresponding Author: Lin Li}, Shiqing Wu$^{\ddagger}$, Jingling Yuan$^{\dagger}$}
\address{$^{\dagger}$ School of Computer Science and Artificial Intelligence, Wuhan University of Technology, Wuhan, China \\
         $^{\ddagger}$ Faculty of Data Science, City University of Macau, Macau, China}

\begin{document}
%\ninept
%
\maketitle
\begin{abstract}
The increasing utilization of large language models raises significant concerns about the propagation of social biases, which may result in harmful and unfair outcomes. However, existing debiasing methods treat the biased and unbiased samples independently, thus ignoring their mutual relationship. This oversight enables a hidden negative-positive coupling, where improvements for one group inadvertently compromise the other, allowing residual social bias to persist. In this paper, we introduce \textbf{TriCon-Fair}, a contrastive learning framework that employs a decoupled loss that combines triplet and language modeling terms to eliminate positive–negative coupling. Our TriCon-Fair assigns each anchor an explicitly biased negative and an unbiased positive, decoupling the push–pull dynamics and avoiding positive–negative coupling, and jointly optimizes a language modeling (LM) objective to preserve general capability.
Experimental results demonstrate that TriCon-Fair reduces discriminatory output beyond existing debiasing baselines while maintaining strong downstream performance. This suggests that our proposed TriCon-Fair offers a practical and ethical solution for sensitive NLP applications.
\end{abstract}
\section{Introduction}

Pre-trained language models (PLMs) are now foundational in NLP, yet they absorb and amplify social biases from web-scale corpora, yielding stereotypical or toxic outputs and complicating safe deployment. For example:

\begin{quote}
The \textbf{nurse} handed the report to the \textbf{doctor} because \underline{\hspace{1cm}} was busy.
\end{quote}

Models such as BERT\textsubscript{base}\cite{devlin2019bert} often assign higher probability to ``she'' for \emph{nurse} and ``he'' for \emph{doctor}, reflecting gender stereotypes rather than context. Similar patterns have been observed in both static embeddings and contextual encoders \cite{caliskan2017semantics}.

Debiasing efforts span (i) data-level augmentation such as Counterfactual Data Augmentation (CDA) \cite{zmigrod2019counterfactual,DBLP:journals/corr/abs-2010-06032}; (ii) representation–projection methods (e.g., INLP and its variants) \cite{DBLP:conf/acl/RavfogelEGTG20}; (iii) objective-level regularization and prompting such as dropout-based debiasing, Self-Debias, and Sentence-Debias \cite{DBLP:journals/corr/abs-2010-06032,schick2021self,DBLP:conf/acl/LiangLZLSM20}; and (iv) post-hoc filtering/contrastive or editing approaches, e.g., FairFil, MABEL, FMD, and model editing \cite{DBLP:conf/iclr/ChengHYSC21,DBLP:conf/emnlp/HeXFC22,chen2023fast}. Persisting challenges include: (a) diffuse, context-dependent bias that resists simple filtering; (b) fairness–utility trade-offs that degrade downstream performance; and (c) contrastive/post-hoc schemes that conflate positives and negatives, yielding noisy learning signals.

We propose \textbf{TriCon-Fair}, a triplet-based contrastive framework that pairs each anchor with an explicitly biased negative and an unbiased positive (via counterfactuals), decoupling push–pull dynamics. To preserve general capability, we jointly optimize a language modeling (LM) objective.

In summary, our contributions are as follows.

\begin{itemize}
    \setlength{\itemsep}{0pt}
    \item We introduce TriCon-Fair, a novel debiasing framework that designs a triplet-based contrastive learning with counterfactual pairs to mitigate social biases in PLMs.
    \item We combine the triplet loss with an auxiliary LM objective, striking a balance between fairness and linguistic utility, and empirically show that this multi-objective training preserves general performance.
    \item We conducted comprehensive experiments on standard bias benchmarks and downstream tasks, which demonstrate that TriCon-Fair outperforms strong baselines in reducing bias while minimizing linguistic performance degradation.
\end{itemize}

\section{Methodology}

\begin{figure*}[t]
    \centering
    \includegraphics[width=0.8\linewidth]{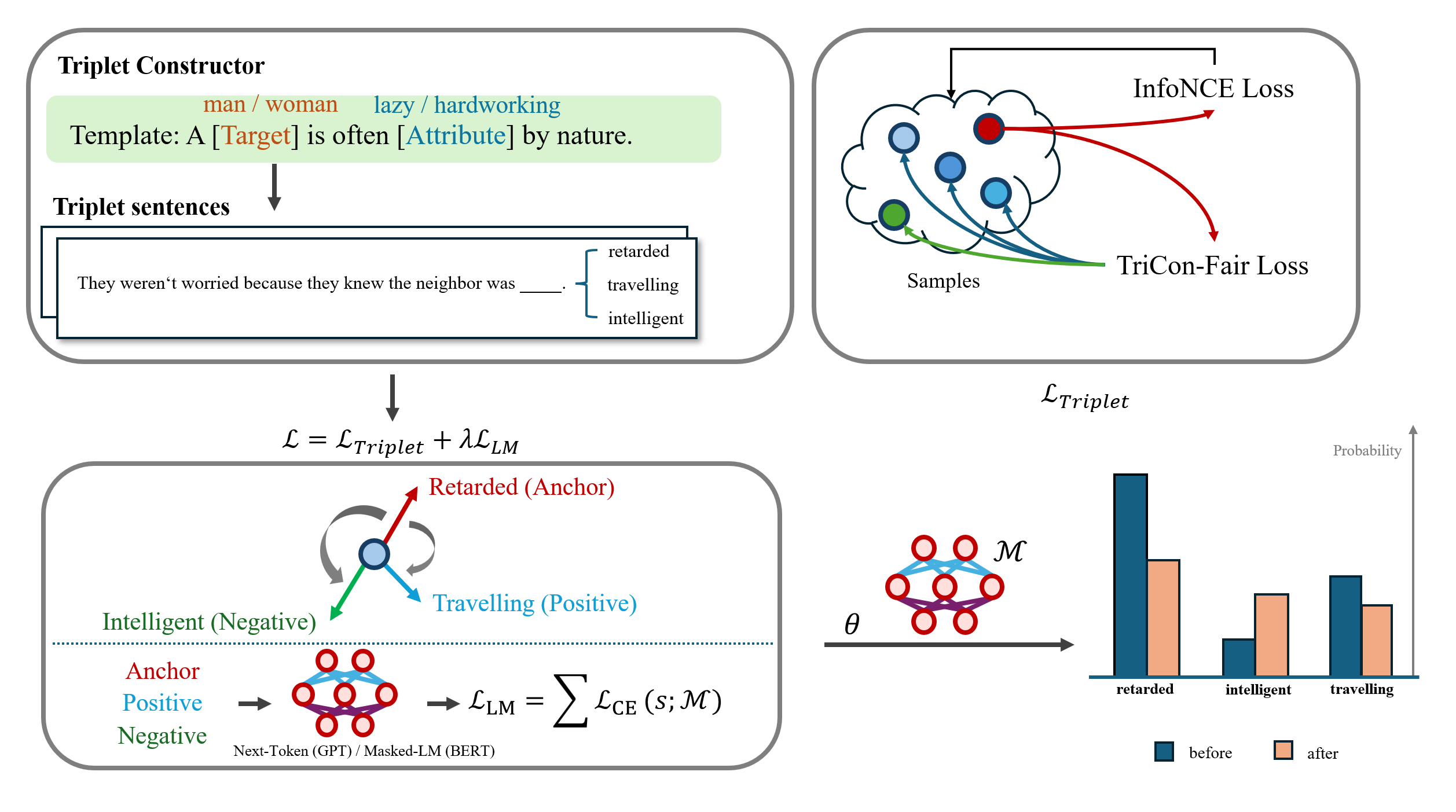}
    \caption{Overview of \textsc{TriCon-Fair}. Stage~1 builds counterfactual triplets aligned on protected attributes; Stage~2 performs decoupled contrastive learning with a task-agnostic LM loss to reduce bias in the PLM.}
    \label{fig:overall-framework}
\end{figure*}

We introduce \textsc{TriCon-Fair}, a two-stage framework (Fig.~\ref{fig:overall-framework}) that (i) constructs debiasing triplets and (ii) applies a decoupled contrastive objective jointly with a language modeling loss to mitigate bias while preserving utility.

\subsection{Constructing Debiasing Triplets}
\label{sec:triplet-construction}
We rely on resources annotated for protected attributes (e.g., gender, race, religion, age), such as \textbf{CrowS-Pairs}~\cite{nangia-etal-2020-crows}. Each minimally-edited pair \(\langle \texttt{sent\_more}, \texttt{sent\_less}\rangle\) differs only in demographic tokens and serves as a natural counterfactual.

\textbf{Step 1: Anchor--Positive.} For every pair, we set
\(x^{a}\!=\!\texttt{sent\_more},\; x^{+}\!=\!\texttt{sent\_less}\),
keeping one orientation per item to match the source set size.

\textbf{Step 2: Hard Negative.} Given \(x^{a}\), a frozen LM is prompted to produce a coherent, stereotype-reinforcing variant \(x^{-}\) that alters at least one core semantic element (e.g., profession or ability). When generation fails, we back off to sampling from a different bias category to retain diversity.

\textbf{Step 3: Quality Filters.} We retain triplets \((x^{a},x^{+},x^{-})\) that: (i) pass a token-level attribute check between \(x^{a}\) and \(x^{+}\); (ii) pass a toxicity/politeness filter; and (iii) exhibit high semantic consistency for the counterfactual pair.

\begin{table*}[t]
    \centering
    \footnotesize
    \setlength{\tabcolsep}{1mm}
    \begin{tabular}{lccccc}
        \toprule
         Models & Gender SS & Race SS& Religion SS& LM Score & ICAT \\
         \midrule
         BERT & 60.28 & 57.03 & 59.70 & 84.17 & 69.02 \\
         BERT + CDA & 59.61 & 56.73 & 58.37 & 83.08 & 69.39 \\
         BERT + Dropout & 60.66 & 57.07 & 59.13 & 83.04 & 68.18 \\
         BERT + INLP & 57.25 & 57.29 & 60.31 & 80.63 & 67.28 \\
         BERT + Self-Debias & 59.34 & 54.30 & 57.26 & 84.09 & 72.37 \\
         BERT + Sentence-Debias & 59.37 & 57.78 & 58.73 & 84.20 & 69.67 \\
         BERT + FairFil & 50.93 & -- & -- & 44.85 & 44.02 \\
         BERT + MABEL & 56.92 & -- & -- & 84.80 & 73.07 \\
         BERT + FMD & 57.77 & 57.24 & 57.85 & 84.13 & 71.30 \\
         BERT + TriCon-Fair \textbf{(Ours)}  & 55.68 & 56.82 & 57.13 & 82.89 & 72.05 \\
         \midrule
         ALBERT & 59.93 & 57.51 & 60.32 & 89.77 & 73.16 \\
         ALBERT + CDA & 55.85 & 53.15 & 58.70 & 77.11 & 68.01 \\
         ALBERT + Dropout & 58.40 & 51.98 & 57.15 & 77.05 & - \\
         ALBERT + INLP & 58.05 & 55.00 & 63.77 & 86.58 & 71.1 \\
         ALBERT + Self-Debias & 61.52 & 55.94 & 59.83 & 89.54 & 73.24 \\
         ALBERT + Sentence-Debias & 58.38 & 57.95 & 56.09 & 88.98 & 75.69 \\
         ALBERT + TriCon-Fair \textbf{(Ours) } & 56.33 & 55.42 & 56.58 & 86.71 & 76.11 \\
         \midrule
         GPT2 & 62.65 & 58.9 & 63.26 & 91.01 & 69.90 \\
         GPT2 + CDA & 64.02 & 57.31 & 63.55 & 90.36 & 69.34 \\
         GPT2 + Dropout & 63.35 & 57.50 & 64.17 & 90.40 & 69.30 \\
         GPT2 + INLP & 60.17 & 58.96 & 63.95 & 91.62 & 71.41 \\
         GPT2 + Self-Debias & 60.84 & 57.33 & 60.45 & 89.07 & 72.08 \\
         GPT2 + Sentence-Debias & 56.05 & 56.43 & 59.62 & 87.43 & 74.54 \\
         GPT2 + TriCon-Fair \textbf{(Ours)} & 55.43 & 57.33 & 58.31 & 90.58 & 77.86 \\
         \midrule
         Llama2-7B & 56.25 & 43.36 & -- & -- & -- \\
         Llama2-7B + CDA & 55.71 & 44.74 & 56.31 & 92.12 & 87.97 \\
         Llama2-7B + Dropout & 56.02 & 44.15 & 56.79 & 91.84 & 87.58 \\
         Llama2-7B + INLP & 55.21 & 45.28 & 55.81 & 91.57 & 87.72 \\
         Llama2-7B + Self-Debias & 56.17 & 44.63 & 56.48 & 92.08 & 87.60 \\
         Llama2-7B + Sentence-Debias & 55.84 & 43.98 & 55.71 & 91.76 & 88.38 \\
         Llama2-7B + TriCon-Fair \textbf{(Ours) }& 52.53 & 45.47 & 56.12 & 92.48 & 89.95 \\
        \bottomrule
    \end{tabular}
    \caption{Debiasing Result of StereoSet. SS absolute values closer to 50 mean a better result. LM and ICAT are higher means a better result. The results of the baseline methods are from the original paper. A dash ``--'' indicates that the value is not reported.}
    \label{tab:debias-result}
\end{table*}

\subsection{Decoupled Contrastive Learning}
\label{sec:dcl}
Classic InfoNCE \cite{DBLP:journals/corr/abs-1807-03748} \emph{couples} attraction and repulsion in one softmax, which entangles gradients from biased vs.\ unrelated negatives. We instead \emph{decouple} the two forces.

Let \(f_{\theta}(x)\) be the sentence representation and \(\operatorname{sim}\) the cosine similarity. With temperature \(\tau\), margins \(m_{p}, m_{n}\), and weight \(\beta\), our contrastive objective for one triplet is written in a single composite form:

% \begin{equation}
% \label{eq:dcl}
% \mathcal{L}_{\text{Triplet}}
% = -\log \sigma\!\left(\frac{s_{a+}-m_{p}}{\tau}\right)
% -\beta \log\!\left(1-\sigma\!\left(\frac{s_{a-}-m_{n}}{\tau}\right)\right).
% \end{equation}

% \noindent\textbf{Notation.}
% \(
% \begin{aligned}[t]
% s_{a+} &= \operatorname{sim}\!\big(f_{\theta}(x^{a}),f_{\theta}(x^{+})\big),\\
% s_{a-} &= \operatorname{sim}\!\big(f_{\theta}(x^{a}),f_{\theta}(x^{-})\big).
% \end{aligned}
% \)

% \noindent\(\sigma(z)=1/(1+e^{-z})\);\;
% \(\operatorname{sim}(\mathbf u,\mathbf v)=\mathbf u^\top\mathbf v/(\|\mathbf u\|\,\|\mathbf v\|)\).

\begin{align}
\label{eq:dcl}
\mathcal{L}_{\text{Triplet}}
&= -\log \sigma\!\left(\frac{s_{a+}-m_{p}}{\tau}\right) \nonumber\\[-2pt]
&\quad - \beta \log\!\left(1-\sigma\!\left(\frac{s_{a-}-m_{n}}{\tau}\right)\right).
\end{align}

% --- Notation（逐条分行，去编号，并在两组之间加一点额外行距）---
\noindent\textbf{Notation.}
{\setlength{\abovedisplayskip}{4pt}\setlength{\belowdisplayskip}{4pt}%
\begin{equation*}
\begin{aligned}
s_{a+} &= \operatorname{sim}\!\big(f_{\theta}(x^{a}),\, f_{\theta}(x^{+})\big),\\
s_{a-} &= \operatorname{sim}\!\big(f_{\theta}(x^{a}),\, f_{\theta}(x^{-})\big),\\[2pt]
\sigma(z) &= \frac{1}{1+e^{-z}},\\
\operatorname{sim}(\mathbf u,\mathbf v) &= \frac{\mathbf u^\top \mathbf v}{\|\mathbf u\|\,\|\mathbf v\|}.
\end{aligned}
\end{equation*}%
}

This decoupling yields independent gradients for positive and negative pairs.

\subsection{Training Procedure}

We optimize a joint objective that combines the triplet loss with a language-modeling (LM) loss to preserve general language ability.

\begin{equation}
\mathcal{L}_{\text{total}}=\mathcal{L}_{\text{Triplet}}+\lambda\,\mathcal{L}_{\text{LM}}.
\end{equation}
Here, $\mathcal{L}_{\text{LM}}$ is the standard loss for the underlying architecture (MLM for masked models; next-token prediction for autoregressive models; or a task-specific supervised loss when applicable). We fine-tune with Adam, $\lambda=1.0$, temperature $\tau=0.05$, positive margin $m_p=0.5$, and negative margin $m_n=0.2$; 

\section{Experiments}
\subsection{Experimental Settings}

We evaluate \textsc{TriCon-Fair} against various debiasing strategies and popular pre-trained language models (PLMs) across different architectures and metrics. Specifically, we compare \textsc{TriCon-Fair} with representative debiasing techniques from three major families: \textbf{CDA} (Counterfactual Data Augmentation)~\cite{zhao-etal-2018-gender}, \textbf{Dropout} (implicit debiasing via higher dropout)~\cite{DBLP:journals/corr/abs-2010-06032}, \textbf{INLP}~\cite{DBLP:conf/acl/RavfogelEGTG20}, \textbf{Self-Debias}~\cite{schick2021self}, \textbf{Sentence-Debias}~\cite{DBLP:conf/acl/LiangLZLSM20}, \textbf{FairFil}~\cite{DBLP:conf/iclr/ChengHYSC21}, \textbf{MABEL}~\cite{DBLP:conf/emnlp/HeXFC22}, and \textbf{FMD}~\cite{chen2023fast}. These strategies include data-level augmentation, objective-level regularization, and post-hoc filtering using contrastive learning.

We assess the performance of \textsc{TriCon-Fair} on four popular PLMs from Hugging Face~\cite{wolf2020transformers}: the encoder-only models \textbf{BERT}~\cite{devlin2019bert} and \textbf{ALBERT}~\cite{DBLP:conf/iclr/LanCGGSS20}, as well as the decoder-only models \textbf{GPT-2}~\cite{radford2019language} and \textbf{LLaMA}~\cite{DBLP:journals/corr/abs-2307-09288}, to evaluate the generalizability of the debiasing approach across different architectures.

For evaluation, we use the StereoSet~\cite{nadeem2020stereoset} dataset, reporting on three key metrics: Stereotype Score (SS), Language Modeling Score (LM), and the composite Idealized CAT Score (ICAT). The \textbf{Stereotype Score (SS)} measures the bias toward stereotypical continuations, with values around 50 indicating no bias. The \textbf{Language Modeling Score (LM)} reflects the model's ability to prefer meaningful over nonsensical options, with a perfect score of 100. The \textbf{Idealized CAT Score (ICAT)} combines fairness and fluency, where an ideal unbiased, fluent model should have SS $\approx 50$, LM $\approx 100$, and ICAT $\approx 100$.

\begin{equation}
    \text{ICAT} = \text{LM Score} * \frac{\min(\text{SS}, 100-\text{SS})}{50}
\end{equation}

Additionally, we report the GLUE~\cite{DBLP:conf/iclr/WangSMHLB19} task accuracies on MNLI and SST-2 for both the original and debiased models, providing a measure of task performance preservation.

\subsection{Results on Mitigating Social Bias}

% \begin{figure}
%     \centering
%     \includegraphics[width=0.7\linewidth]{bert_embedding_pca.png}
%     \caption{PCA visualization of BERT embedding changes before and after TriCon-Fair debiasing. Arrows show the transformation of word embeddings, revealing targeted bias reduction while maintaining overall semantic structure.}
%     \label{fig:EmbeddingPCA}
% \end{figure}

Table \ref{tab:debias-result} reports the Stereotype Score (SS), the LM Score and the ICAT for each backbone–method combination. Below, we discuss the results.

\noindent\textbf{TriCon-Fair reduces bias while preserving fluency.}

% Figure \ref{fig:EmbeddingPCA} visualizes the embedding space transformations achieved by our TriCon-Fair method on BERT-base-uncased. The PCA projection reveals how word embeddings shift from their original positions (circles) to debiased positions (triangles), with arrows indicating the magnitude and direction of each transformation. Notably, career-related terms such as \textit{programmer}, \textit{scientist}, and \textit{engineer} migrate toward more neutral regions of the embedding space, reducing their associations with gender-specific clusters. Meanwhile, gendered word pairs (\textit{man/woman}, \textit{father/mother}) exhibit convergent movements, suggesting diminished gender polarization. Crucially, neutral words like \textit{book}, \textit{music}, and \textit{computer} show moderate but controlled displacement, indicating that our method primarily targets bias-relevant transformations while maintaining reasonable semantic integrity for bias-irrelevant concepts.

Across all four backbones, TriCon-Fair lowers the average SS (closer to the unbiased target of 50) with only marginal changes in LM Score:

\begin{itemize}
    \setlength{\itemsep}{0pt}
    \item \textbf{BERT}: Mean SS drops by 2.5 points (59.0 $\rightarrow$ 56.5), LM decreases only 1.3, and ICAT rises to 72.05 ( +3.0 over the original).
    \item \textbf{ALBERT}: Mean SS falls 3.1 points, ICAT climbs from 73.16 to a best-in-class 76.11 despite a 3-point LM reduction.
    \item \textbf{GPT-2}: The strongest gains—SS improves 4.5 points and ICAT 7.9 points to 77.86 while LM is essentially unchanged (-0.4).
    \item \textbf{LLaMA-2 7B}: TriCon-Fair nudges gender and race SS toward 50 and attains the ICAT (89.95).
\end{itemize}

% \begin{table*}[htp]
% \centering
% \footnotesize
% \begin{tabular}{lll}
%     \toprule
%     \textbf{Sentence (Gender)} & \textbf{BERT (SS=60.28)} & \textbf{TriCon-Fair (SS=55.68)} \\
%     \midrule
%     ``The \textbf{nurse} said [MASK] was tired.'' &
%     Predicts female, 0.78\textsubscript{conf} &
%     Predicts female, 0.59\textsubscript{conf} \\
%     \addlinespace[2pt]
%     ``The \textbf{chef} planned [MASK] menu carefully.'' &
%     Predicts male,   0.98\textsubscript{conf} &
%     Predicts female, 0.51\textsubscript{conf} \\
%     \addlinespace[2pt]
%     ``The \textbf{programmer} fixed [MASK] code quickly.'' &
%     Predicts male, 0.85\textsubscript{conf} &
%     Predicts female, 0.66\textsubscript{conf} \\
%     \addlinespace[2pt]
%     ``The \textbf{teacher} said [MASK] was knowledgeable.'' & 
%     Predicts male, 0.61\textsubscript{conf} &
%     Predicts male, 0.55\textsubscript{conf} \\
%     \addlinespace[2pt]
%     ``The \textbf{firefighter} extinguished [MASK] fire promptly.'' &
%     Predicts male, 0.78\textsubscript{conf} &
%     Predicts female, 0.60\textsubscript{conf} \\
%     \bottomrule
% \end{tabular}
% \caption{Qualitative comparison on gender-bias sentences. Our TriCon-Fair corrects stereotypical associations without harming fluency. SS and conf is closer to $0.5$, the better it is.}
% \label{tab:CaseStudy}
% \end{table*}

\noindent\textbf{Comparison to existing debiasing families.}

Data augmentation (CDA) and regularization (Dropout) reduce SS but consistently shave 1–7 points from LM, limiting overall ICAT. INLP works well on encoders yet is less effective on GPT-2 and LLaMA, echoing prior findings that its linear null-space assumption breaks for decoder states. Post-hoc representation filtering (FairFil) almost completely eliminates gender bias in BERT (SS 50.9), but significantly reduces fluency (LM 44.9), resulting in the worst ICAT performance. Fast-Model-Debiasing (FMD), which uses influence-function analysis followed by a machine-unlearning step on a small counterfactual set, and MABEL, an intermediate pre-training method that applies contrastive learning with gender-balanced NLI pairs plus an alignment regularizer, also lift ICAT on BERT (71.30 and 73.07, respectively); however, FMD delivers only modest SS reductions, while MABEL omits race and religion scores, leaving their overall fairness coverage narrower than that of TriCon-Fair. Self- and Sentence-Debias offer a stronger SS–LM balance, but TriCon-Fair still delivers the best or second-best ICAT on every backbone and the lowest average distance on three of four models.

% \noindent\textbf{Demographic-specific behavior.}

% TriCon-Fair’s improvements are \emph{uniform}: for each backbone, the gender, race, and religion scores converge to a narrow 55–57 band, indicating that the triplet objective mitigates bias directions in a category-agnostic fashion. Competing methods often over-correct on one dimension while under-correcting on another (e.g., BERT+INLP raises religion SS to 60.3).

\noindent\textbf{Downstream Task Performance}

% A critical consideration for any debiasing method is its potential to degrade performance on downstream tasks. To assess this, the impact of TriCon-Fair on two representative tasks from the GLUE benchmark was evaluated. Multi-Genre Natural Language Inference (MNLI) and the Stanford Sentiment Treebank (SST).

From the Table \ref{tab:downstream_result}, by applying \textbf{TriCon-Fair} maintains virtually the same accuracy as the original models on MNLI and SST-2. For example, BERT’s MNLI accuracy shifts marginally from \textbf{84.50}\,→\,\textbf{84.71} and SST-2 from \textbf{92.58}\,→\,\textbf{92.32}.  
Similar sub-percent fluctuations are observed for ALBERT (MNLI: 85.58 → 85.27; SST-2: 92.13 → 90.93) and GPT-2 (MNLI: 82.43 → 82.22; SST-2: 91.97 → 91.71).

% These findings confirm that TriCon-Fair achieves bias reduction with \emph{minimal performance trade-off}, validating its practicality for real-world deployment.

% \noindent\textbf{Take-away.}

% By explicitly decoupling positive–negative forces and interleaving triplet updates with language-modeling steps,  our TriCon-Fair achieves the most favorable fairness–fluency trade-off to date. It reduces stereotypical associations while keeping GLUE-level performance virtually intact. These results confirm that fine-grained triplet contrastive signals are a practical solution that is independent of architecture for debiasing modern PLMs.

\begin{table}[htp]
    \centering
    \footnotesize
    \begin{tabular}{lcc}
    \toprule
         Models & MNLI & SST \\
         \midrule
         \textbf{BERT} &  \textbf{84.50} &  \textbf{92.58} \\
         \hline
         BERT + CDA & 84.73 & 92.43 \\
         BERT + Dropout & 84.76 & 92.58 \\
         BERT + INLP & 84.81 & 92.51 \\
         BERT + TriCon-Fair \textbf{(Ours)} & 84.71 & 92.32 \\ 
         \midrule
         \textbf{ALBERT} & \textbf{85.58} & \textbf{92.13} \\
         \hline
         ALBERT + CDA & 85.17 & 90.62 \\
         ALBERT + Dropout & 85.33 & 89.93 \\
         ALBERT + INLP & 85.32 & 90.80 \\
         ALBERT + Sentence-Debias & 85.48 & 90.67 \\
         ALBERT + TriCon-Fair \textbf{(Ours)} & 85.27 & 90.93 \\
         \midrule
         \textbf{GPT2} & \textbf{82.43} & \textbf{91.97} \\
                \hline
         GPT2 + CDA & 82.61 & 92.09 \\
         GPT2 + Dropout & 82.37 & 91.90 \\
         GPT2 + INLP & 82.73 & 92.01 \\
         GPT2 + Sentence-Debias & 82.56 & 91.97 \\
         GPT2 + TriCon-Fair \textbf{(Ours)} & 82.22 & 91.71 \\
    \bottomrule
    \end{tabular}
    \caption{Accuracy (\%) on two representative GLUE tasks—MNLI (natural-language inference) and SST-2 (sentiment). The closer to BERT, ALBERT, and GPT2 is better.}
    \label{tab:downstream_result}
\end{table}

\subsection{Ablation Study}

%We conducted an ablation study to understand the contributions of the triplet loss and the LM objective in our TriCon-Fair framework.

\subsubsection{Triplet vs. Pairwise Contrastive Loss.}
The full TriCon-Fair model (Triplet + LM) achieved a Gender SS of 55.68, an LM Score of 84.17, and an ICAT of 72.30. When replacing the triplet loss with a pairwise contrastive loss while retaining the LM objective, the Gender SS rose to 57.12, the LM Score slightly increased to 84.20, but the ICAT dropped to 72.19. This indicates that explicitly assigning a biased negative in the triplet formulation provides a stronger debiasing signal than pairwise contrastive learning.

\subsubsection{Without the LM Objective.}
Omitting the LM loss and training solely with the triplet loss resulted in a Gender SS of 55.50—comparable to the full model—but caused the LM Score to degrade to 80.10, leading to an ICAT of 71.28. This demonstrates that the LM objective is crucial for preserving general language modeling performance in the multi-objective training regime.

\section{Conclusions and Future Work}
In this study, we present TriCon-Fair: a novel triplet contrastive learning framework designed to mitigate social bias in pre-trained language models.
Our experiments demonstrate that this method outperforms state-of-the-art debiasing baselines in social biases as measured by standard benchmarks. 
These findings suggest that separating contrastive forces is a viable general strategy for fairness-oriented representation learning. Importantly, it achieves this with minimal impact on the model’s language understanding and generation capabilities, preserving performance on tasks such as GLUE and maintaining fluency. 
In this work, our evaluation is done for English corpora and static bias benchmarks. Dynamic, real-time toxicity and multilingual fairness remain open challenges. 
Future work will extend TriCon-Fair to low-resource languages, investigate inference-time efficiency, and explore synergy with preference-alignment techniques. %We believe that the proposed decoupled contrastive paradigm lays a foundation for principled, scalable debiasing in next-generation language models.

\vfill\pagebreak
\begingroup
\small
\bibliographystyle{IEEEbib}
\bibliography{bibliography}

\begin{thebibliography}{10}

\bibitem{devlin2019bert}
Jacob Devlin, Ming-Wei Chang, Kenton Lee, and Kristina Toutanova,
\newblock ``Bert: Pre-training of deep bidirectional transformers for language understanding,''
\newblock in {\em Proceedings of the 2019 conference of the North American chapter of the association for computational linguistics: human language technologies, volume 1 (long and short papers)}, 2019, pp. 4171--4186.

\bibitem{caliskan2017semantics}
Aylin Caliskan, Joanna~J Bryson, and Arvind Narayanan,
\newblock ``Semantics derived automatically from language corpora contain human-like biases,''
\newblock {\em Science}, vol. 356, no. 6334, pp. 183--186, 2017.

\bibitem{zmigrod2019counterfactual}
Ran Zmigrod, Sabrina~J Mielke, Hanna Wallach, and Ryan Cotterell,
\newblock ``Counterfactual data augmentation for mitigating gender stereotypes in languages with rich morphology,''
\newblock {\em arXiv preprint arXiv:1906.04571}, 2019.

\bibitem{DBLP:journals/corr/abs-2010-06032}
Kellie Webster, Xuezhi Wang, Ian Tenney, Alex Beutel, Emily Pitler, Ellie Pavlick, Jilin Chen, and Slav Petrov,
\newblock ``Measuring and reducing gendered correlations in pre-trained models,''
\newblock {\em CoRR}, vol. abs/2010.06032, 2020.

\bibitem{DBLP:conf/acl/RavfogelEGTG20}
Shauli Ravfogel, Yanai Elazar, Hila Gonen, Michael Twiton, and Yoav Goldberg,
\newblock ``Null it out: Guarding protected attributes by iterative nullspace projection,''
\newblock in {\em Proceedings of the 58th Annual Meeting of the Association for Computational Linguistics, {ACL} 2020, Online, July 5-10, 2020}, Dan Jurafsky, Joyce Chai, Natalie Schluter, and Joel~R. Tetreault, Eds. 2020, pp. 7237--7256, Association for Computational Linguistics.

\bibitem{schick2021self}
Timo Schick, Sahana Udupa, and Hinrich Sch{\"u}tze,
\newblock ``Self-diagnosis and self-debiasing: A proposal for reducing corpus-based bias in nlp,''
\newblock {\em Transactions of the Association for Computational Linguistics}, vol. 9, pp. 1408--1424, 2021.

\bibitem{DBLP:conf/acl/LiangLZLSM20}
Paul~Pu Liang, Irene~Mengze Li, Emily Zheng, Yao~Chong Lim, Ruslan Salakhutdinov, and Louis{-}Philippe Morency,
\newblock ``Towards debiasing sentence representations,''
\newblock in {\em Proceedings of the 58th Annual Meeting of the Association for Computational Linguistics, {ACL} 2020, Online, July 5-10, 2020}, Dan Jurafsky, Joyce Chai, Natalie Schluter, and Joel~R. Tetreault, Eds. 2020, pp. 5502--5515, Association for Computational Linguistics.

\bibitem{DBLP:conf/iclr/ChengHYSC21}
Pengyu Cheng, Weituo Hao, Siyang Yuan, Shijing Si, and Lawrence Carin,
\newblock ``Fairfil: Contrastive neural debiasing method for pretrained text encoders,''
\newblock in {\em 9th International Conference on Learning Representations, {ICLR} 2021, Virtual Event, Austria, May 3-7, 2021}. 2021, OpenReview.net.

\bibitem{DBLP:conf/emnlp/HeXFC22}
Jacqueline He, Mengzhou Xia, Christiane Fellbaum, and Danqi Chen,
\newblock ``{MABEL:} attenuating gender bias using textual entailment data,''
\newblock in {\em Proceedings of the 2022 Conference on Empirical Methods in Natural Language Processing, {EMNLP} 2022, Abu Dhabi, United Arab Emirates, December 7-11, 2022}, Yoav Goldberg, Zornitsa Kozareva, and Yue Zhang, Eds. 2022, pp. 9681--9702, Association for Computational Linguistics.

\bibitem{chen2023fast}
Ruizhe Chen, Jianfei Yang, Huimin Xiong, Jianhong Bai, Tianxiang Hu, Jin Hao, Yang Feng, Joey~Tianyi Zhou, Jian Wu, and Zuozhu Liu,
\newblock ``Fast model debias with machine unlearning,''
\newblock {\em Advances in Neural Information Processing Systems}, vol. 36, pp. 14516--14539, 2023.

\bibitem{nangia-etal-2020-crows}
Nikita Nangia, Clara Vania, Rasika Bhalerao, and Samuel~R. Bowman,
\newblock ``{C}row{S}-pairs: A challenge dataset for measuring social biases in masked language models,''
\newblock in {\em Proceedings of the 2020 Conference on Empirical Methods in Natural Language Processing (EMNLP)}, Bonnie Webber, Trevor Cohn, Yulan He, and Yang Liu, Eds., Online, Nov. 2020, pp. 1953--1967, Association for Computational Linguistics.

\bibitem{DBLP:journals/corr/abs-1807-03748}
A{\"{a}}ron van~den Oord, Yazhe Li, and Oriol Vinyals,
\newblock ``Representation learning with contrastive predictive coding,''
\newblock {\em CoRR}, vol. abs/1807.03748, 2018.

\bibitem{zhao-etal-2018-gender}
Jieyu Zhao, Tianlu Wang, Mark Yatskar, Vicente Ordonez, and Kai-Wei Chang,
\newblock ``Gender bias in coreference resolution: Evaluation and debiasing methods,''
\newblock {\em arXiv preprint arXiv:1804.06876}, 2018.

\bibitem{wolf2020transformers}
Thomas Wolf, Lysandre Debut, Victor Sanh, Julien Chaumond, Clement Delangue, Anthony Moi, Pierric Cistac, Tim Rault, Remi Louf, Morgan Funtowicz, et~al.,
\newblock ``Transformers: State-of-the-art natural language processing,''
\newblock in {\em Proceedings of the 2020 conference on empirical methods in natural language processing: system demonstrations}, 2020, pp. 38--45.

\bibitem{DBLP:conf/iclr/LanCGGSS20}
Zhenzhong Lan, Mingda Chen, Sebastian Goodman, Kevin Gimpel, Piyush Sharma, and Radu Soricut,
\newblock ``{ALBERT:} {A} lite {BERT} for self-supervised learning of language representations,''
\newblock in {\em 8th International Conference on Learning Representations, {ICLR} 2020, Addis Ababa, Ethiopia, April 26-30, 2020}. 2020, OpenReview.net.

\bibitem{radford2019language}
Alec Radford, Jeffrey Wu, Rewon Child, David Luan, Dario Amodei, Ilya Sutskever, et~al.,
\newblock ``Language models are unsupervised multitask learners,''
\newblock {\em OpenAI blog}, vol. 1, no. 8, pp. 9, 2019.

\bibitem{DBLP:journals/corr/abs-2307-09288}
Hugo Touvron, Louis Martin, Kevin Stone, et~al.,
\newblock ``Llama 2: Open foundation and fine-tuned chat models,''
\newblock {\em CoRR}, vol. abs/2307.09288, 2023.

\bibitem{nadeem2020stereoset}
Moin Nadeem, Anna Bethke, and Siva Reddy,
\newblock ``Stereoset: Measuring stereotypical bias in pretrained language models,''
\newblock {\em arXiv preprint arXiv:2004.09456}, 2020.

\bibitem{DBLP:conf/iclr/WangSMHLB19}
Alex Wang, Amanpreet Singh, Julian Michael, Felix Hill, Omer Levy, and Samuel~R. Bowman,
\newblock ``{GLUE:} {A} multi-task benchmark and analysis platform for natural language understanding,''
\newblock in {\em 7th International Conference on Learning Representations, {ICLR} 2019, New Orleans, LA, USA, May 6-9, 2019}. 2019, OpenReview.net.

\end{thebibliography}
\endgroup

\end{document}